\documentclass{ecai} 

\usepackage{amsmath}
\usepackage{graphicx}
\usepackage{algorithm}
\usepackage{algpseudocode}
\usepackage{multirow}
\usepackage{hyperref}

\newcommand{\BibTeX}{B\kern-.05em{\sc i\kern-.025em b}\kern-.08em\TeX}

\begin{document}


\begin{frontmatter}


\paperid{1420} 


\title{Fair-OBNC: Correcting Label Noise for Fairer Datasets}


\author[A]{\fnms{Inês}~\snm{Oliveira e Silva}
\thanks{Corresponding Author. Email: ines.silva@feedzai.com}}
\author[A, C]{\fnms{Sérgio}~\snm{Jesus}
}
\author[A]{\fnms{Hugo}~\snm{Ferreira}
} 
\author[A]{\fnms{Pedro}~\snm{Saleiro}
}
\author[B]{\fnms{Inês}~\snm{Sousa}
}
\author[A]{\fnms{Pedro}~\snm{Bizarro}
}
\author[C]{\fnms{Carlos}~\snm{Soares}
}

\address[A]{Feedzai}
\address[B]{Fraunhofer Portugal AICOS}
\address[C]{Universidade do Porto}


\begin{abstract}
Data used by automated decision-making systems, such as Machine Learning models, often reflects discriminatory behavior that occurred in the past. These biases in the training data are sometimes related to label noise, such as in COMPAS, where more African-American offenders are wrongly labeled as having a higher risk of recidivism when compared to their White counterparts. Models trained on such biased data may perpetuate or even aggravate the biases with respect to sensitive information, such as gender, race, or age. However, while multiple label noise correction approaches are available in the literature, these focus on model performance exclusively. 
In this work, we propose Fair-OBNC, a label noise correction method with fairness considerations, to produce training datasets with measurable demographic parity. The presented method adapts Ordering-Based Noise Correction, with an adjusted criterion of ordering, based both on the margin of error of an ensemble, and the potential increase in the observed demographic parity of the dataset. 
We evaluate Fair-OBNC against other different pre-processing techniques, under different scenarios of controlled label noise. 
Our results show that the proposed method is the overall better alternative within the pool of label correction methods, being capable of attaining better reconstructions of the original labels. Models trained in the corrected data have an increase, on average, of 150\% in demographic parity, when compared to models trained in data with noisy labels, across the considered levels of label noise.
\end{abstract}

\end{frontmatter}


\section{Introduction}
\label{sec:introduction}

The widespread usage of Machine Learning (ML) systems in sensitive use cases may impact people's lives profoundly when given the ability to make high-stakes decisions~\cite{mehrabi2021survey}. One well-known example is the Correctional Offender Management Profiling for Alternative Sanctions (COMPAS) software. This software assesses the recidivism risk of individuals, and a score produced by it is used by American courts to decide whether a person should be released from prison with bail. In a 2016 investigation conducted by ProPublica,\footnote{https://www.propublica.org/article/machine-bias-risk-assessments-in-criminal-sentencing} it was discovered that the system was biased against African-Americans, incorrectly labeling Black offenders as ``high-risk'' twice as often as White offenders. Many other biased AI examples exist, under different settings and use-cases~\cite{dastin2022amazon,datta2014automated,obermeyer2019dissecting,sweeney2013discrimination,zarsky2014understanding}. As observed by ProPublica, we can classify this specific algorithm as unfair, as its decisions reflect a form of discrimination or preference, towards certain groups of people based on their inherent or acquired characteristics~\cite{mehrabi2021survey}. The general goal of \emph{Fair ML} is to identify and mitigate harmful inequalities observed in many of these predictive systems~\cite{barocas-hardt-narayanan}. 

When collecting data, discrimination is bound to lead to incorrect labels, and the relationship between the sensitive information of an instance and its assigned class will be biased. This is especially problematic in settings with the selective labels problem~\cite{lakkaraju2017selective}, where the true labels of a given instance are not obtainable, depending on the decision it was made. In these scenarios, it is common practice to propagate the decision as the true label for future iterations of the dataset. Going back to the previous COMPAS example, it is not possible to determine if an offender would re-offend or not if bail is denied, and it is assumed that a positive decision (denying bail) is a positive label (recidivism) for the dataset. Learning to predict the outcome of a certain task involves generalizing from historical examples, which can lead to algorithms perpetuating or even exacerbating the biases present in the labels of past decisions and observations~\cite{zhao2022towards}. However, despite the vast amount of literature on methods for dealing with noisy data, only a few of these studies focus on identifying and correcting noisy labels~\cite{nicholson2015label}. While it is intuitive to think about label noise correction techniques being applied to eliminate past discrimination, there is still a gap in developing methods that aim to improve the fairness of datasets used to train ML models through label noise correction. The existing correction methods focus on predictive performance of the resulting models~\cite{feng2015class, nicholson2015label, sun2007identifying, tanaka2018joint, xu2020hybrid, zheng2020error}.

In this work, we address this gap by proposing Fair-OBNC, a novel method for fairness-aware label correction. This method is an extension of Ordering-Based Label Noise Correction (OBNC)~\cite{feng2015class}, which additionally uses the information provided by the sensitive attributes and class prevalence when correcting instances with noisy labels. We propose three major modifications to the original method: a) modify the ordering criterion which determines labels to be corrected, b) ignore the sensitive attribute and/or proxy variables in the process of determining the probability of each label being noisy, and c) allow the user to select either to use a committee voting scheme or a score-based margin. 
More specifically, regarding the ranking of noisy instances, instead of correcting a fixed number of the most likely labels, our method additionally focuses on achieving a certain ceiling of disparity of class prevalence between the sensitive groups (\textit{i.e.}, Demographic Parity), filtering the ordered instances to correct based on this requirement.
We empirically evaluate this method against multiple pre-processing baselines on a realistic fraud detection dataset, to which we inject controlled label noise scenarios in different ways. We control the noise within the dataset by setting a probability of changing the labels of instances belonging to a specific sensitive group and class. We evaluate the methods at increasing noise rates, and observe the changes in both performance and fairness. We additionally present metrics for the correctness of label correction for methods that alter the labels of the original noisy dataset.

Our experiments show that, within the pool of label noise correction methods, Fair-OBNC consistently presents the best results in both reconstruction and fairness of the data after correction in the criterion of Demographic Parity. Additionally, when compared to other pre-processing Fair ML techniques, Fair-OBNC produces dominant results in terms of fairness, albeit with lower performance than some approaches, for the levels of noise tested. 

To summarize, we make several contributions to the field of fairness in machine learning, particularly in the context of label noise correction:

\begin{itemize}
    \item We present a novel fairness-aware label noise correction method, extending the OBNC algorithm with fairness considerations. This method specifically targets the increase of demographic parity in training datasets, making it a unique approach in the intersection of fairness and noise correction.
    \item We develop an extensive experimental setup on a fraud detection dataset with controlled and biased noise injection, simulating multiple noise scenarios. 
    \item We provide a comparative analysis of the performance of Fair-OBNC against several baseline methods, conducting a benchmark of multiple preprocessing-based fairness-enhancing techniques. This comprehensive evaluation demonstrates the advantages of our approach in terms of fairness metrics and label reconstruction accuracy.
\end{itemize}

\section{Related work}
\label{sec:relatedwork}

In this section, we present the relevant literature regarding label noise, existing methods to perform label noise correction, and strategies to ensure ML fairness.

\subsection{Label noise}

\textit{Noise} can be defined as non-systematic errors that possibly degrade an algorithm's ability to learn the relationship between the features and the true label of a sample~\cite{frenay2013classification}. When noise is related to defective labels, we are in the presence of label noise. Label noise is a common phenomenon in real-world datasets, and the cost of acquiring non-polluted data is usually high~\cite{algan2021image}.

Label noise is particularly important in the case of bias mitigation techniques, as data bias and label corruption are closely related. This happens because the accuracy of the labels is often affected by the subject belonging to a protected group~\cite{wang2021fair}. However, bias mitigation techniques and metrics typically assume that the provided labels are correct. Allied to this fact, it is possible that a segment of decisions will produce actions that render the true labels unattainable. This is exemplified by Eq.~\ref{eq:selective_labels} for binary classification problems, where the true label for negative decisions is undetermined. 

\begin{equation}
\label{eq:selective_labels}
  y_i=
  \begin{cases}
    0 \text{ or } 1, & \text{if } \hat{y}_i = 1 \\
    \emptyset, & \text{Otherwise}    
  \end{cases}
\end{equation}

This is defined as the Selective Labels problem~\cite{lakkaraju2017selective}. Common approaches for this problem include using the decision $\hat{y}_i$ as the label $y_i$ for the next dataset, or removing samples where the label can not be determined. In the presence of biased decisions, both options can further increase disparities in future models.

Label noise can be classified into one of the following categories~\cite{algan2021image}. Let $x$ denote the feature vector of a given instance, $s$ represent the sensitive attribute or attributes within the feature vector, $y$ indicate its clean label and $y^*$ denote the corresponding noisy or observed label. \textbf{Random noise} is randomly distributed and does not depend on the instance's features or label, meaning that all instances share the same probability of having an incorrect label, expressed as $P(y^*|y, x, s) = P(y^*)$. \textbf{Y-dependent noise} happens when instances belonging to a particular class are more likely to be mislabeled and therefore the probability of having a label flipped is dependent on the true label, expressed as $P(y^*|y, x, s) = P(y^*|y)$. Finally, \textbf{XY-dependent noise} depends on both feature values and target values. 
If the noise specifically depends on the sensitive attribute, it is referred to as group-dependent~\cite{wang2021fair} or instance-dependent~\cite{wu2022fair} in the fairness literature and is related to discrimination. In such cases, the probability of an instance being noisy depends both on its class and on the sensitive group it belongs to. Assuming a scenario where the sensitive attribute can have two possible values, $0$ and $1$, this means that the probability of an instance from the positive sensitive group being noisy is different from that of an instance belonging to the negative group, which can be expressed as $P(y^*|y, x, s=0) \neq P(y^*|y, x, s=1)$.

\subsection{Label noise correction}
\label{ssec:}

In this work, we focus on the types of noise-dealing approaches that deal with noise by identifying and correcting the corrupted labels. One strategy for this problem is to analyze the predictions of an ensemble of classifiers trained on bootstraps of the dataset to determine which instances might be incorrectly labeled. The predictions are used to estimate the probability of an instance being noisy~\cite{sun2007identifying, feng2015class} or to assign a new label to an instance using a model-committee voting scheme~\cite{nicholson2015label}. Other approaches include determining first which instances are likely to be correctly labeled, and subsequently using those to learn how to correct the mislabeled instances~\cite{nicholson2015label} or, leveraging clustering methods to identify incorrect labels~\cite{nicholson2015label, xu2020hybrid}.

From these methods, we highlight one which uses ensemble margins as a measure of how likely an instance is to be mislabeled~\cite{feng2015class}, which we name OBNC. This method first trains an ensemble classifier with the originally noisy data. The ensemble predicts a label for each instance, and the ensemble margins are calculated for the mislabeled ones. The authors discuss several ways of calculating the ensemble margins, depending on the number of classes to use in the margin computation (two or all) and if the margin should be computed according to the observed label or most voted class (supervised or unsupervised). In this work, since we are considering a case of supervised binary classification, we chose to focus on the supervised margin definition~\cite{bartlett1998boosting}. This specific ensemble margin, $M$, is defined by Equation~\ref{eq:ensemblemargin}, where $v$ is the total number of votes given by the number of classifiers in the ensemble and $v_y$ is the number of votes for the observed label $y$:
 
\begin{equation}
\label{eq:ensemblemargin}
    M(x,y) = \frac{2 v_y - v}{v} \text{.}
\end{equation}

The margin values take values in the range $[-1,1]$, where misclassified instances by the majority of the classifiers have negative margin values. These samples are ranked in descending order according to a label noise evaluation function that relies on the ensemble's margin, as defined in Equation~\ref{eq:obnc_evalaution}, where $Tr$ is the training set of $(x, y)$ pairs, and $C$ is the ensemble classifier.

\begin{equation}
\label{eq:obnc_evalaution}
    N(x_i, y_i) = |M(x_i, y_i)| \: , \: \forall (x_i, y_i) \in Tr | C(x_i) \neq y_i
\end{equation}

\begin{figure}[ht]
    \centering
    \includegraphics[width=0.45\textwidth]{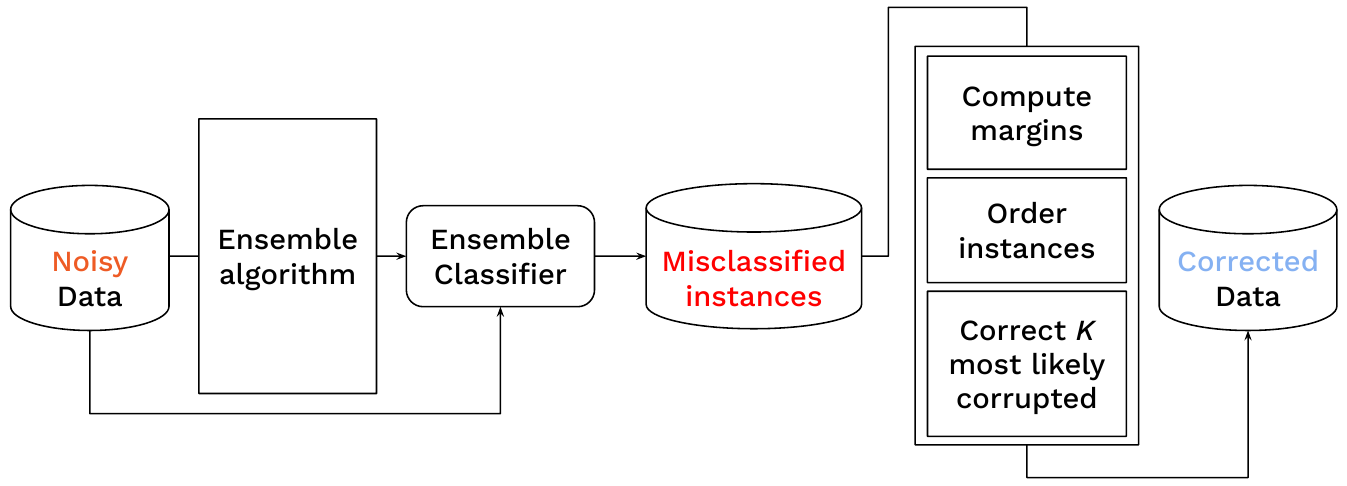}
    \caption{The Ordering-Based Noise Correction method.}
    \label{fig:obnc}
\end{figure}

Higher values of $N(x_i, y_i)$ indicate an estimated higher probability of the observed label in the dataset $y_i$ being noisy. Finally, the $K$ samples with the highest values of $N(x_i, y_i)$ are corrected (with $K$ being a hyperparameter of the method), changing their labels to the ones predicted by the ensemble. A diagram depicting the described process for OBNC is shown in Figure~\ref{fig:obnc}.

\subsection{Fairness-enhancing Mechanisms}

Several methods have been proposed for tackling the problem of Fair ML. The existing literature commonly divides these mechanisms into three categories, according to which part of the ML pipeline they focus on: pre-processing, in-processing, and post-processing mechanisms~\cite{d2017conscientious,mehrabi2021survey, pessach2022review}. We will be focusing on pre-processing methods, which generally transform the input data for fairer outcomes, as is the case of the introduced method. In contrast, in-processing methods focus on ensuring fairness during the training process, adapting algorithms by including regularization, reweighting, or constraining the optimization to account for fairness~\cite{d2017conscientious, mehrabi2021survey, pessach2022review, lamba2021empirical}. Post-processing methods apply some form of transformation to the predictions or scores in order to remove biases from the results~\cite{mehrabi2021survey,pessach2022review}.

\subsubsection{Pre-processing methods}

Pre-processing approaches aim to eliminate bias by modifying data before training a model, some focusing on changing or reweighting the labels, and others on altering feature representations~\cite{d2017conscientious, pessach2022review}. 

Resampling techniques adjust the distribution of the training data so that disparities in class balance are minimized between the groups defined by the sensitive attribute or attributes~\cite{calders2010three, iosifidis2019fae}. Some methods work by altering the labels of instances (by their closeness to the decision boundary of a ranker~\cite{kamiran2012data}, or to achieve a certain desired feature distribution~\cite{feldman2015certifying}, for example), while others change the representativity of data through re-weighting the dataset, depending on the class and group of each instance~\cite{kamiran2012data, jiang2020identifying}. As an alternative to re-weighting and relabeling, a dataset may also be resampled to achieve the same non-discriminatory label distribution~\cite{kamiran2012data}. Other different lines of work include creating implicitly balanced datasets for learning algorithms, so that demographic parity and equality of opportunity are achieved~\cite{kehrenberg2020tuning}, as well as the technique of increasing-to-balance~\cite{liu2021can}, which consists of inserting noise to balance noise rates and fairness metrics across classes. 

On the other hand, fairness can be enhanced by altering feature representations. One approach is to alter the data by using a convex optimization method that transforms the biased data with a randomized mapping function that balances the trade-offs between increasing fairness, while constraining sample distortion and  utility for downstream models~\cite{calmon2017optimized}. 
A different approach to obtain fair representations of data takes human inputs on instance similarity as additional information in creating a fairness graph~\cite{lahoti2019operationalizing}. The authors combine the data-driven similarity with the supplementary pairwise information from the graph to learn Pairwise Fair Representations.

Our technique is a novel pre-processing method that corrects labels according to their likelihood of being noisy, introducing fairness criteria in the correction process to simultaneously balance group prevalence disparities.

\section{Fair Ordering-Based Noise Correction}
\label{sec:fairobnc}

With the objective of reducing the disparities present in the labels through label noise correction, we propose Fair-OBNC, a variation of the OBNC method that additionally optimizes for fairness.

We consider the notion of demographic parity~\cite{dwork2012fairness}, which is reached when the probability of a decision $\hat{y}$ is independent of the sensitive attribute $s$, \textit{i.e.}, $P(\hat{y}|s) = P(\hat{y})$. For the Fair-OBNC algorithm, instead of the decision, we are aiming for the labels after correction $y_{\mathcal{R}}$. 
The correction algorithm is modified to select the instances that are more likely to be wrongly labeled, and improve demographic parity of the observed labels in train $y^*$. The process describing how this criterion is applied is shown in Algorithm~\ref{alg:fairordering}.

\begin{algorithm}[ht]
\caption{Fair-OBNC}\label{alg:fairordering}
    \begin{algorithmic}[1]
        \Require ordered misclassified instances' index vector $I$, margin scores vector $S$, maximum label flipping rate $R$, disparity target $D$, margin threshold $T$, sensitive attribute vector $A$, corresponding label vector $Y$
        \State $K \gets R \cdot |Y|$
        \State $c \gets 0$
        \State $F_0 \gets F(0)$ \Comment{Number of instances of group 0 to flip}
        \State $F_1 \gets F(1)$ \Comment{Number of instances of group 1 to flip}
        \For{$i \in I$}
            \If{$(S[i] < T) \lor (c = K)$}
                \State \text{break} \Comment{Stopping criteria was met}
            \EndIf
            \State $s \gets A[i]$
            \If{$(F_s > 0 \land Y[i] = 0) \lor(F_s < 0 \land Y[i] = 1)$}
                \State $Y[i] \gets 1 - Y[i]$ \Comment{Flip instance if it reduces disparity}
                \State $c \gets c + 1$
                \If{$F_s > 0$} \Comment{Update number of instances to flip}
                    \State $F_s \gets F_s - 1$
                \Else
                    \State $F_s \gets F_s + 1$
                \EndIf
            \EndIf
        \EndFor
    \end{algorithmic}
\end{algorithm}

An important step of our algorithm is calculating the number of instances of each group (defined by the sensitive attribute) that are required to change their label, in order to achieve a desired target disparity $D$. To do so, we determine the range of prevalence values that each sensitive group must have to achieve a prevalence disparity below $D$. These minimum ($P(y)_l$) and maximum ($P(y)_h$) prevalence values are defined in Eq.~\ref{eq:min_prevalence} and~\ref{eq:max_prevalence}, respectively.

\begin{equation}
    \label{eq:min_prevalence}
    P(y)_l = P(y) \left ( 1-D \right )
\end{equation}
\begin{equation}
    \label{eq:max_prevalence}
    P(y)_h = P(y) \left ( 1+D \right )
\end{equation}

The number of instances to flip, $F(s)$, can then be calculated for each sensitive group $s$, as defined in Eq.~\ref{eq:nflips}, where $S$ represents the instances belonging to the sensitive group $s$.
\begin{equation}
    \label{eq:nflips}
    F(s) = 
        \begin{cases}
            |S|(P(y|s)-P(y)_{h}), & \text{if }P(y|s) > P(y)_{h} \\ 
            |S|(P(y)_{l}-P(y|s)), & \text{if }P(y|s) < P(y)_{l} \\
            0, & \text{Otherwise}
        \end{cases}
\end{equation}

This number is positive if the group prevalence is below the overall prevalence; this indicates that it is beneficial to flip observed negative samples of that group to have a higher number of final positive labels. On the other hand, a negative value indicates that the group prevalence is above the total prevalence; conversely, it is beneficial to flip positive samples. 

After ordering the misclassified labels by decreasing margins, we assess whether performing the correction will be beneficial in achieving a balanced distribution of labels. Each instance is flipped if it belongs to the positive class and a group with a higher prevalence than the upper limit defined by the prevalence disparity, or if it belongs to the negative class and a group with lower prevalence than the lower limit. 

Additionally, instead of correcting the $K$ instances with higher margin values, we instead define a maximum flip rate $R$, and further consider a margin threshold $T$. As such, when iterating through the ordered instances, we stop correcting if either we have corrected $R$ of the instances, the margin score of the instances is lower than $T$, or the demographic parity of all groups is above $D$.

A different approach for fair ML is to remove the sensitive attribute from the training data~\cite{kamiran2012data}. This is done expecting the decisions of the resulting model to be independent of sensitive attributes when this information is removed. This correction is situational, as other features may be correlated to the sensitive attribute, and perpetuate the biases observed in models trained with the sensitive information~\cite{lamba2021empirical}. We allow ignoring the sensitive attributes, as well as any other potentially correlated features when training the ensemble of classifiers and calculating the margins.

The last modification to the original algorithm is the inclusion of a parameter to determine the margin calculation. These can be calculated with the predicted class, in a voting system, replicating the original algorithm, or with the predicted score given by the $n$ classifiers of the ensemble $\phi$. In this case, the margin is defined as Eq.~\ref{eq:residuals_margin}.

\begin{equation}
    \label{eq:residuals_margin}
    M(x, y) = -\left|y - \frac{1}{n}\sum^{n}_{i=1}\phi_i(x)\right|
\end{equation}

With this change, we increase the resolution of the margin vector, as the score vector is constituted with float numbers, when compared to binary class predictions. This consequently reduces the number of margin ties in the dataset.

In summary, our method, Fair-OBNC, combines three main modifications: performing the corrections to decrease the disparity in the prevalence of positive labels between both groups, ignoring the sensitive attributes when training the ensemble, and allowing to calculate the margins based on the scores rather than predicted label of the models in the ensemble.

\section{Experiments}
\label{sec:experiments}

This section explains the experimental setup used for evaluating our proposed method for fair label noise correction. We also characterize the used datasets and describe the methodology applied for the systematic injection of label noise. The necessary code to replicate the conducted experiments, as well as the supplementary material, can be found at \url{https://github.com/feedzai/fair-obnc.}

\subsection{Baseline Methods}

We chose to conduct our experiments on a range of pre-processing methods that either modify the labels, the features, or the data distribution, to understand how our method fares in comparison to different types of approaches. For all the considered methods and FairOBNC (named LabelFlipping in the package\footnotemark[\value{footnote}]), we used the implementation available in  Aequitas~\cite{jesus2024aequitas}. We consider the following baselines: 

\begin{itemize}
    \item \textbf{No pre-processing};
    \item The original \textbf{OBNC} method~\cite{feng2015class};
    \item \textbf{Prevalence Sampling}~\cite{lamba2021empirical}, a random sampling method that aims to equalize class prevalence for the groups in the dataset;
    \item \textbf{Data Repairer}~\cite{feldman2015certifying}, a method that transforms the data distribution so that a given feature distribution is independent of the sensitive attributes. This is achieved by matching the conditional distribution $P(X|s)$ to the global variable distribution $P(X)$;
    \item \textbf{Massaging}~\cite{kamiran2012data}, which consists of changing the labels of instances depending on the group, selected by a singular model (\textit{i.e.}, ranker);
    \item \textbf{Suppression}~\cite{kamiran2012data}, where the objective is to find and remove the most related features to the sensitive attribute. We do this in two ways: 
    \begin{itemize}
        \item Correlation-based, meaning that all attributes with a correlation value with the sensitive attribute above a certain tuneable threshold are removed;
        \item Feature Importance-based, by training a classifier to predict the sensitive attribute value based on the features and removing the feature with the highest learned importance. This is performed iteratively, performing backward feature selection until no feature exhibits any significant relation to the sensitive attribute.
    \end{itemize}
\end{itemize}

\footnotetext{\url{https://github.com/dssg/aequitas}}
\subsection{Dataset}

Throughout the conducted experiments, we use the \textit{Bank Account Fraud} dataset~\cite{jesus2022turning}, namely the \textit{Variant II} of the suite. This is a binary classification tabular dataset for bank account opening fraud detection. The dataset is anonymized, with the applicant's age being the sensitive attribute available, which is also a binary variable. The first sensitive group includes all applicants whose age is equal to or above 50 ($s=A$), while the remaining applicants who are less than 50 years old belong to the second group ($s=B$). We select this variant of the suite because the groups defined by the sensitive attribute have approximately equal sizes. We balance the different label prevalence depending on group prior to the noise injection step.

\subsection{Label noise injection}

To systematically assess the effectiveness of our method against the considered baselines under diverse biased data scenarios, we inject label noise in multiple ways to unbiased versions of the described dataset~\cite{silva2023systematic}. 

The first step of this process involves generating an independent and identically distributed (IID) dataset, in regard to the sensitive attribute and the data splits. This is done by first shuffling the sensitive attribute column, which consequently removes any existing relationship between this column, the label, and features. Then, instances are shuffled randomly to train, validation, and test sets, to remove any potential data drift in the original splits. To confirm these properties, classifiers were trained to predict both the sensitive attribute and the split. The resulting models were observed to be random predictors. 

The subsequent label noise injection depends on the label, and on the sensitive group each instance belongs to. Regarding the label, we consider three scenarios: applying label noise to both labels equally, only applying noise to the sample with positive labels, and only applying it to the sample with negative labels. Considering the sensitive group, we uniformly apply noise to instances of one of the groups (group $A$ only), simulating a scenario where the sensitive attribute determines the probability of an incorrect label. With these scenarios, we observe how the considered methods perform under different levels of noise.

\subsection{Hyperparameters}

For each pre-processing method, we randomly sample 50 hyperparameter configurations and apply them to the training set. The hyperparameter spaces where the samples are created from are defined in the supplementary material, in Section B.1.  The modified train sets obtained from the different pre-processing methods are used to train LightGBM models, a state-of-the-art algorithm for tabular data~\cite{shi2022quantized}, with hyperparameters equally sampled from a grid. These models make predictions for a corresponding clean test set. Because of the label noise scenario, metrics (presented in Section \ref{sec:metrics}) are averaged by the 50 runs of each method. This is because under label noise, it might not be possible to determine the best model for a task, since that a model selected by the best metrics in a validation set might not actually be the best performing model in non-noisy data. 

\subsection{Metrics}
\label{sec:metrics}

\begin{table*}[ht]
    \caption{Label noise correction performance across the considered noise rates (averaged over 50 trials with different hyperparameters obtained by random search) for label noise injected in each label separately and simultaneously.}
  \begin{tabular}{ll|ccc|ccc|ccc}
    \hline 
    \multicolumn{2}{c|}{} &
      \multicolumn{3}{c|}{\textbf{Noise rate: 5\%}} &
      \multicolumn{3}{c|}{\textbf{Noise rate: 10\%}} &
      \multicolumn{3}{c}{\textbf{Noise rate: 20\%}} \\
    \multicolumn{2}{c|}{} & OBNC & Fair-OBNC & Massaging & OBNC & Fair-OBNC & Massaging & OBNC & Fair-OBNC & Massaging \\
    \hline
    \multirow{7}{*}{\rotatebox[origin=c]{90}{Label 0}} & Reconstruction Score & 0.752 & \textbf{0.979} & 0.974 & 0.773 & \textbf{0.961} & 0.949 & 0.813 & \textbf{0.924} & 0.899 \\
    & FPR & 0.240 & \textbf{0.020} & 0.026 & 0.219 & \textbf{0.039} & 0.051 & 0.179 & \textbf{0.076} & 0.101 \\
    & FNR & 0.949 & 0.102 & \textbf{0.029} & 0.937 & 0.114 & \textbf{0.031} & 0.909 & 0.120 & \textbf{0.031} \\
    & FPR (group B) & 0.246 & \textbf{0.011} & 0.025 & 0.228 & \textbf{0.020} & 0.050 & 0.190 & \textbf{0.038} & 0.101 \\
    & FNR (group B) & 0.946 & \textbf{0.000} & \textbf{0.000} & 0.934 & \textbf{0.000} & \textbf{0.000} & 0.903 & \textbf{0.000} & \textbf{0.000} \\
    & FPR (group A) & 0.234 & 0.029 & \textbf{0.026} & 0.210 & 0.057 & \textbf{0.051} & 0.169 & 0.113 & \textbf{0.102}\\
    & FNR (group A) & 0.952 & 0.204 & \textbf{0.057} & 0.940 & 0.227 & \textbf{0.061} & 0.916 & 0.239 & \textbf{0.062} \\ \hline
    \multirow{7}{*}{\rotatebox[origin=c]{90}{Label 1}} & Reconstruction Score & 0.729 & \textbf{0.999} & \textbf{0.999} & 0.729 & \textbf{0.999} & \textbf{0.999} & 0.729 & \textbf{0.998} & \textbf{0.998} \\ 
    & FPR & 0.263 & \textbf{0.000} & \textbf{0.000} & 0.263 & \textbf{0.000} & \textbf{0.000} & 0.264 & \textbf{0.000} & 0.001 \\ 
    & FNR & 0.978 & \textbf{0.025} & 0.042 & 0.970 & \textbf{0.056} & 0.079 & 0.954 & \textbf{0.129} & 0.151 \\ 
    & FPR (group B) & 0.263 & \textbf{0.000} & \textbf{0.000} & 0.263 & \textbf{0.000} & \textbf{0.000} & 0.264 & \textbf{0.000} & \textbf{0.000} \\
    & FNR (group B) & 0.987 & \textbf{0.000} & 0.034 & 0.987 & \textbf{0.012} & 0.059 & 0.988 & \textbf{0.064} & 0.108 \\
    & FPR (group A) & 0.263 & 0.001 & \textbf{0.000} & 0.263 & \textbf{0.000} & 0.001 & 0.263 & \textbf{0.000} & 0.001 \\ 
    & FNR (group A) & 0.969 & \textbf{0.049} & 0.050 & 0.953 & 0.099 & \textbf{0.098} & 0.920 & 0.194 & \textbf{0.193} \\ \hline
    \multirow{7}{*}{\rotatebox[origin=c]{90}{Both Labels}} & Reconstruction Score & 0.752 & \textbf{0.979} & 0.974 & 0.773 & \textbf{0.960} & 0.949 & 0.812 & \textbf{0.923} & 0.898 \\ 
    & FPR & 0.240 & \textbf{0.020} & 0.026 & 0.219 & \textbf{0.039} & 0.051 & 0.180 & \textbf{0.076} & 0.101 \\ 
    & FNR & 0.943 & 0.122 & \textbf{0.051} & 0.927 & 0.154 & \textbf{0.078} & 0.897 & 0.199 & \textbf{0.134} \\ 
    & FPR (group B) & 0.246 & \textbf{0.011} & 0.025 & 0.229 & \textbf{0.020} & 0.050 & 0.191 & \textbf{0.037} & 0.100 \\ 
    & FNR (group B) & 0.946 & \textbf{0.000} & \textbf{0.000} & 0.934 & \textbf{0.000} & \textbf{0.000} & 0.903 & \textbf{0.000} & \textbf{0.000} \\ 
    & FPR (group A) & 0.235 & 0.030 & \textbf{0.026} & 0.210 & 0.057 & \textbf{0.052} & 0.169 & 0.113 & \textbf{0.103} \\ 
    & FNR (group A) & 0.941 & 0.243 & \textbf{0.103} & 0.920 & 0.307 & \textbf{0.155} & 0.890 & 0.398 & \textbf{0.267} \\ \hline
  \end{tabular}
  
\label{tab:reconstruction_scores}
\end{table*}

The task of the dataset is detection of fraud in account opening, and the main objective is to identify as many fraudulent attempts as possible. Fraudulent instances are labeled as positive, and as such, the predictive performance of the models is measured by the True Positive Rate ($TPR$). This metric is defined in Eq.~\ref{eq:tpr}, where $TP$ represents true positives and $FN$ denotes false negatives.  

\begin{equation}
\label{eq:tpr}
    TPR = \frac{TP}{TP + FN}
\end{equation}

A threshold must be defined, to limit the number of false positive predictions. To do so, we select the value for which 1\% of the highest model scores (\textit{i.e.}, the top 1\% of scores are classified as fraudulent). We select this threshold value as it is the approximate prevalence of fraud in the clean data.

To empirically evaluate whether ML models are discriminatory, we measure the Demographic Parity of the predictions~\cite{dwork2012fairness}, as defined in Eq.~\ref{eq:demographicparity}.

\begin{equation}
\label{eq:demographicparity}
    P(\hat{y}=1|s=A)=P(\hat{y}=1|s=B)
\end{equation}

We calculate the ratio between the lowest and highest predicted prevalence to obtain a metric of Demographic Parity. In this metric, values close to $1$ represent an equilibrium of predicted prevalence between every group in the dataset, while values close to $0$ represent higher disparities in the fairness metric.

Particularly considering the label noise correction methods, we additionally want to measure their efficacy at identifying and correcting instances with incorrect labels. To do so, we compare the transformed training labels (\textit{i.e.}, after applying the label noise correction) to the original clean labels from the IID dataset, measuring each method's reconstruction abilities by calculating a \emph{Reconstruction Score}. This is defined in Equation~\ref{eq:similarity}, where $y_i$ represents a clean label from a given instance $i$ and ${y_\mathcal{R}}_i$ the label value after the processes of label noise injection and correction for the same instance $i$. 

\begin{equation}
    \text{\emph{Reconstruction Score}} = \frac{\sum_{i=1}^N {y_\mathcal{R}}_i = y_i}{N}
    \label{eq:similarity}
\end{equation}

We further calculate the false positive and false negative rates ($FPR_\mathcal{R}$ and $FNR_\mathcal{R}$, respectively) following the same logic of comparison between the clean and corrected labels. Here, we consider the false positive rate to be the ratio of instances whose label was negative in the clean data but was modified to positive (either by noise injection or by the noise correction algorithm). This is shown in Eq.~\ref{eq:fpr_r}.

\begin{equation}
\label{eq:fpr_r}
    FPR_\mathcal{R} = \frac{FP_\mathcal{R}}{FP_\mathcal{R} + TN_\mathcal{R}}
\end{equation}

Following the same logic, the false negative rate corresponds to the proportion of clean positive labels that are negative after the noise correction was applied, as presented in Eq.~\ref{eq:fnr_r}.

\begin{equation}
\label{eq:fnr_r}
    FNR_\mathcal{R} = \frac{FN_\mathcal{R}}{FN_\mathcal{R} + TP_\mathcal{R}}
\end{equation}

\section{Results}
\label{sec:results}

We examine the results of the conducted experiments in this section. We analyze both the reconstruction abilities of the label noise correction methods, and the downstream metrics for models trained in all pre-processing methods in terms of both predictive performance and fairness.

\subsection{Evaluation of label correction}
\label{sec:labelcorrectionevaluation}

Our first analysis aims at better understanding how well each label noise correction method fares at correctly identifying label noise. We evaluate the \emph{Reconstruction Score}, $FPR_\mathcal{R}$ and $FNR_\mathcal{R}$, as described in Section~\ref{sec:metrics}. We present the false positive and negative rates globally, as well as for each sensitive group separately. The values presented in Table~\ref{tab:reconstruction_scores} were obtained by averaging the results of each method for each noise rate over the 50 trials with different hyperparameters obtained by random search.

\subsubsection{Injecting noise on the negative label}

By analyzing the "Label 0" rows of Table~\ref{tab:reconstruction_scores}, we can observe that Fair-OBNC achieves the highest reconstruction score, closely followed by Massaging. The original OBNC method achieves a significantly lower score when compared to the previous methods. Unlike the other two methods, the reconstruction score obtained by OBNC increases with a higher noise rate. This happens due to the stopping criteria for the OBNC method, which is exclusively a fixed value defined by the user, or alternatively, exhausting instances with negative margins. Since, on average, OBNC corrects around 27\% of the instances, as the noise rate increases, the number of labels being corrected gets closer to the actual noise rate. As a consequence of the noise level approaching the average number of corrected instances, both the number of false positives and negatives decreases for OBNC. 

Regarding FPR and FNR for group $B$, where no noise was injected, we can observe that Fair-OBNC and Massaging have no false negatives. Since the prevalence is lower for this group, both these methods only correct negative instances,  and as a result only have false positives. OBNC, on the other hand, flips the labels of most positive instances in this group and obtains an FNR close to 1. Furthermore, we can see that Fair-OBNC also maintains lower false positive rates than the other methods for this group across all noise rates. 

Finally, for group $A$, Massaging achieves a slightly lower FPR than Fair-OBNC. This represents noisy labels that were not corrected in the presented noise scenario. We additionally observe a significantly lower FNR in Massaging, which represents clean labels that were incorrectly flipped.

\subsubsection{Injecting noise on the positive label}
 
Let us now consider the results obtained when injecting noise on the positive label, which are reported in the "Label 1" rows of Table~\ref{tab:reconstruction_scores}.

Similarly to the previous results, we observe that the OBNC method corrects a large number of instances before the stopping criteria are met. This affects its reconstruction performance, having high false positive and negative rates, when compared to the other two methods.

For the other two methods, despite an almost perfect reconstruction score, it is also worth noting that the number of noisy labels is reduced, as the dataset is imbalanced towards a low number of positive instances. As such, both these methods also correct a low number of instances (less than 1\% of the total dataset). This is visible in the FNR metric of group $A$, which is close to the noise rate, showing how barely any of the injected noisy labels were correctly identified.

\subsubsection{Injecting noise on both labels}

Finally, we focus on the results obtained when injecting noise in both classes, presented in the "Both Labels" rows of Table~\ref{tab:reconstruction_scores}. In this scenario, the methods behave similarly to when only injecting noise in the negative class, to which the majority of the instances belong to. This is justifiable by the low prevalence of positive labels in the dataset, similarly to the previous case.
We can once again observe how the Fair-OBNC method achieves the highest reconstruction scores, for all noise rates.

\subsection{Evaluation of performance}

We evaluate the obtained models on the clean labels of IID test set to investigate how the methods would perform in a testing scenario where the biases that were present in the training data have been corrected in more recent data.

\begin{figure}[ht]
    \centering
    \includegraphics[width=0.45\textwidth]{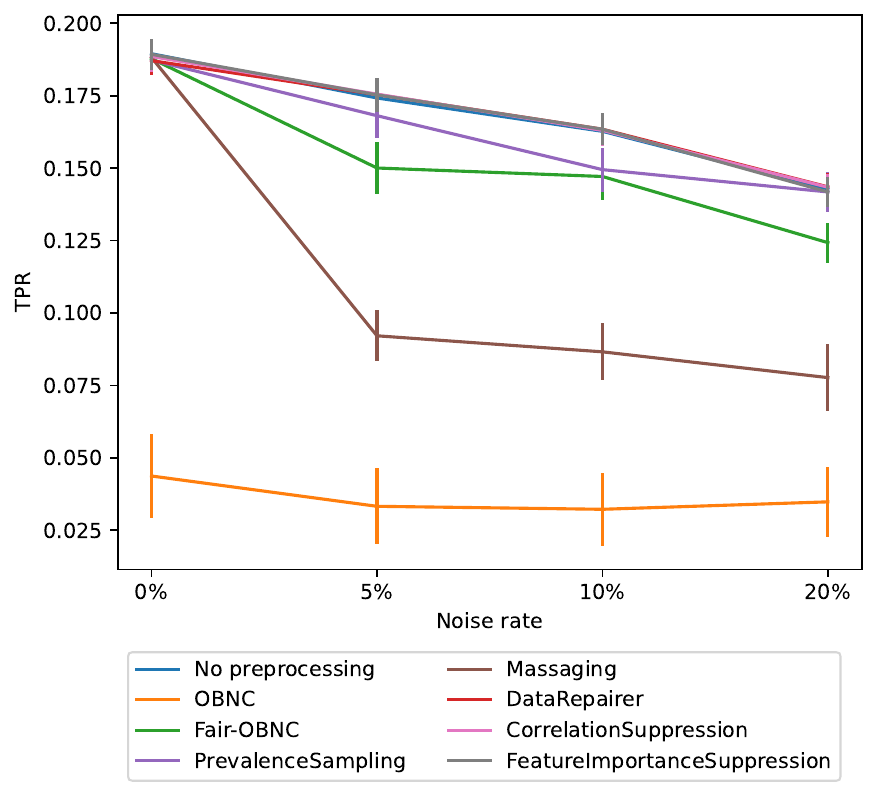}
    \caption{True Positive Rate (averaged over 50 trials) achieved for each pre-processing method, for increasing noise rates.}
    \label{fig:label0_tpr}
\end{figure}

\begin{figure}[ht]
    \centering
    \includegraphics[width=0.45\textwidth]{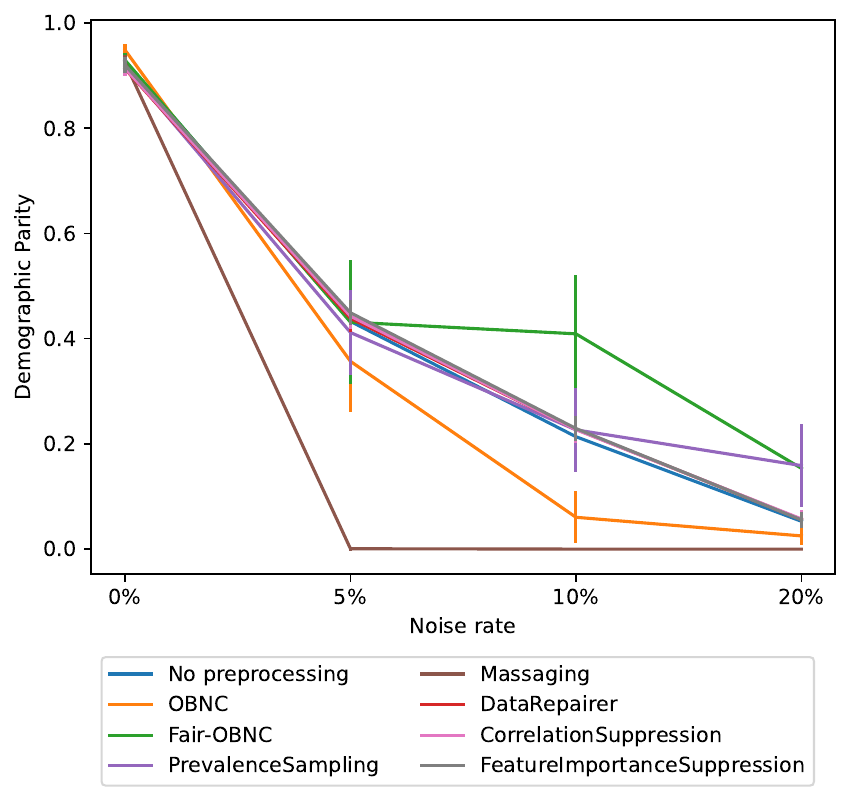}
    \caption{Demographic Parity (averaged over 50 trials) achieved for each pre-processing method, for increasing noise rates.}
    \label{fig:label0_dempar}
\end{figure}

Similarly to the results observed in Section~\ref{sec:labelcorrectionevaluation}, the performance when only injecting noise in the positive label instances barely changes among methods and noise rates, due to the very small number of positive cases (and consequently mislabeled instances) in the dataset. Another consequence of this is the fact that the results for the other two scenarios are very similar, and the same conclusions can be derived from analyzing when injecting noise only on the negative label, and on both labels. For these reasons, we only elaborate on results obtained from injecting noise on the negative label instances. The observed TPR and Demographic Parity across the increasing noise rates can be found in Figs.~\ref{fig:label0_tpr} and~\ref{fig:label0_dempar}. We provide the obtained results for the cases where noise is injected in the positive label and in both labels simultaneously, along with the results of additional experiments conducted using different noise injection processes in the supplementary material, in Section C.

The OBNC method exhibited significantly poorer predictive performance compared to the other methods, with similar fairness levels in the initial levels of noise injection. It is important to note that fairness alone does not indicate a good model, as random classifiers are fair in its predictions, following the definition of demographic parity.

The Massaging method experienced a drastic drop in fairness to zero with any level of noise. This occurred due to the high score attributed by the model to instances that are corrected to positives, in conjunction with the definition of the threshold. As a consequence, only the corrected instances from the group with lower prevalence were classified as positive, and no instances from the group with higher prevalence had a positive prediction. This results in zero demographic parity, as defined by demographic parity's dependence on predicted prevalence.

Fair-OBNC and Prevalence Sampling achieve comparable performance in terms of fairness, with Fair-OBNC showing a  significantly higher Demographic Parity at a noise rate of 10\%. Despite the fairness gain, compared to not applying any pre-processing to the labels, both methods incurred a slight decrease in performance.

The models obtained from the data modified using any of the remainder methods do not show differences to the models obtained using no pre-processing. 

\section{Discussion}
\label{sec:discussion}

In the conducted empirical evaluation, we observe that the proposed modifications to the OBNC method resulted in significant improvements in both fairness and predictive performance when compared to the original technique and other label correction methods. Our method achieves results comparable to the top performing method, Prevalence Sampling, showing the positive impact of modifying existing label noise correction specifically for the task of improving the fairness of ML models. In this analysis, we do not include statistical significance tests. We do believe conclusions would remain similar with that inclusion.


By further analyzing the reconstruction accuracy of the label noise correction methods, we can also conclude that the accuracy of the original OBNC algorithm is highly dependent on properly defining how many labels to correct, which is not possible in most cases since the noise model is unknown. By changing the criteria for stopping the label noise correction, we were able to overcome this significant limitation of the original OBNC method.

Nevertheless, we identify opportunities for improvement in the algorithm we propose. 
Firstly, the method ignores the sensitive attribute. This is a simple strategy that works well in this case where we arbitrarily select a single sensitive attribute and are in control of the noise injection process. However, this strategy may be too simplistic to achieve fairness in more complicated scenarios where there might be multiple sensitive attributes and more complex noise models. It is also known that in many cases it is possible to infer the sensitive information from the attributes (or features) in the dataset~\cite{lamba2021empirical}, having high correlations or even completely different distributions for different groups. In such cases, this adjustment would not be effective.

Furthermore, the intuition behind the application of fairness criteria when deciding which labels to correct could be extended to apply different and more elaborate fairness definitions.

Finally, the algorithm we propose is limited to binary classification problems. The scope of this work could be broadened by adapting this strategy to handle multiclass scenarios. Another interesting direction would be to allow for targeting the repair of labels to a single group, e.g. the minority class, or a single label. This second case is particularly relevant in the context of the selective labels problem~\cite{lakkaraju2017selective}, enabling the identification and correction of labels whose value has been conditioned on previous model decisions.

\section{Conclusions}
\label{sec:conclusions}

In this work, we tackle the problem of learning fair ML classifiers from biased data. We propose a method for fair label noise correction that applies multiple modifications to the Ordering-Based Noise Correction method~\cite{feng2015class} to take fairness into consideration during noise correction. These variations include ignoring the sensitive attribute and deciding which labels to correct in a way that balances the distribution of classes across groups. In the conducted experiments, we evaluate our method against the original OBNC method, as well as multiple pre-processing baselines, and analyzed the fairness and predictive performance of the obtained models. We observed that the modifications introduced to an existing label noise correction method resulted in models that performed better and were less biased in predicting fraudulent cases when compared to the original approach. Our method revealed superior performance to most of the fairness-enhancing methods considered in the empirical evaluation, achieving results comparable to the Prevalence Sampling technique.




\begin{ack}
This work was partially funded by projects Agenda “Center for Responsible AI”, nr. C645008882-00000055, investment project nr. 62, financed by the Recovery and Resilience Plan (PRR) and by European Union -  NextGeneration EU and FCT plurianual funding for 2020-2023 of LIACC (UIDB/00027/2020 UIDP/00027/2020).
\end{ack}


\bibliography{m1420}

\end{document}